\documentclass{article}
\usepackage[nonatbib, final]{nips_style}
\usepackage[utf8]{inputenc} 
\usepackage[T1]{fontenc}    
\usepackage{hyperref}       
\usepackage{url}            
\usepackage{booktabs}       
\usepackage{amsfonts}       
\usepackage{nicefrac}       
\usepackage{microtype}      
\usepackage{float}
\usepackage{subcaption}
\usepackage{graphicx}
\usepackage{cite}
\usepackage{enumitem}
\usepackage[table,xcdraw]{xcolor}
\usepackage{adjustbox}
\usepackage{hyperref}
\usepackage{svg}
\usepackage{sectsty}
\usepackage{titlesec}
\usepackage{url}
\usepackage{float}
\usepackage{microtype}
\usepackage{graphicx} 
\usepackage{amsmath}
\usepackage[utf8]{inputenc} 
\usepackage[T1]{fontenc}    
\usepackage{hyperref}       
\usepackage{url}            
\usepackage{booktabs}       
\usepackage{amsfonts}       
\usepackage{nicefrac}       
\usepackage{microtype}      
\usepackage{xcolor}         
\usepackage{xspace}
\usepackage{amsmath}
\usepackage{amssymb}
\usepackage{booktabs}
\usepackage{makecell}
\usepackage{colortbl}
\usepackage{graphicx}   
\usepackage{CJKutf8}
\usepackage{multirow}
\usepackage{interval}
\intervalconfig{soft open fences}
\usepackage{algorithm, algpseudocode}

\usepackage{float}
\usepackage{enumitem}
\usepackage{tabularx}
\usepackage{placeins}
\usepackage{color}
\usepackage{tcolorbox}
\usepackage{verbatim}
\usepackage{colortbl} 
\usepackage{listings}
\usepackage{makecell}
\usepackage{siunitx}
\usepackage{xcolor}
\definecolor{bg}{RGB}{176,226,255}
\definecolor{bonus_green}{RGB}{0,100,0}
\newcommand{\bbonus}[1]{{\textcolor{bonus_green}{$^{\uparrow#1}$}}}

\title{One-shot Entropy Minimization}

\author{
\textbf{Zitian Gao} \quad
\textbf{Lynx Chen} \quad
\textbf{Haoming Luo} \quad
\textbf{Joey Zhou} \quad
\textbf{Bryan Dai}\thanks{~Corresponding author.} \vspace{2mm} \\
\hspace{6mm} Ubiquant \quad \vspace{2mm} \\
\texttt{\{ztgao02,ylchen,hmluo,jzhou,cbdai\}@ubiquant.com} \\
}

\begin{document}
\maketitle

\begin{abstract}
We trained 13,440 large language models and found that entropy minimization requires only a single unlabeled data and 10 steps optimization to achieve performance improvements greater than those obtained using thousands of data and carefully designed rewards in rule-based reinforcement learning. This striking result may prompt a rethinking of post-training paradigms for large language models. Our code is avaliable at \url{https://github.com/zitian-gao/one-shot-em}.

\end{abstract}

\section{Introduction}

The post-training phase of large language models (LLMs) has advanced rapidly~\cite{xu2025largereasoningmodelssurvey,logicrl,orz,1shot,prime,zoo}, with models like DeepSeek-R1~\cite{deepseekai2025deepseekr1incentivizingreasoningcapability}, Kimi-K1.5~\cite{kimiteam2025kimik15scalingreinforcement}, and OpenAI o-series~\cite{o1,o3} demonstrating remarkable reasoning abilities. However, preparing for Reinforcement Learning (RL) is never an easy task. It often requires a large amount of high-quality ground truth labeled data, along with the careful design of rule-based rewards to maximize advantage signals and prevent reward hacking. In contrast, Entropy Minimization (EM) is entirely unsupervised. We used the EM method to train 13,440 large language models in order to eliminate randomness in training as much as possible and ensure that the experimental results and observed patterns are reliable. Our rigorous study demonstrates that using just a single piece of unlabeled data, the performance already surpasses that of traditional RL. Moreover, they typically converge within just 10 training steps, which is significantly faster than the thousands of steps often required for RL. EM is based on two direct and simple assumptions: (1) The sampling process in generation of large language models is inherently stochastic, (2) Correct answers generally have lower entropy than incorrect ones. Our study reveals that EM and RL share the same goal: unlocking the pretrained model’s latent potential without adding new knowledge\cite{under}. Both rely on a process we call “token reranking” to maximize the model’s performance. We find that entropy minimization has the capacity to rival RL in the post-training phase.

Our main contributions are as follows:
\vspace{-1mm}
\begin{itemize}[leftmargin=*]
    \item We propose \textbf{\textit{One-shot Entropy Minimization}}, a surprisingly powerful and fully unsupervised method that rivals or surpasses reinforcement learning using just a single unlabeled data.

    \item We conduct an in-depth analysis of the effectiveness of One-shot EM, making it a highly reasonable approach. We find that it shares many core properties with RL, yet drives model behavior in the opposite direction when viewed through the lens of \textbf{\textit{logits shift}}. 

    \item We extensively evaluate EM and identify temperature as a key factor for both training and inference, and EM shows an opposite trend to RL with respect to inference-time temperature.

    \item We reveal that EM is a distribution shaping tool rather than a learning method by analyzing the inconsistency in loss-reasoning curve over confidence in Section~\ref{loss_avg}) and the logit shift(in Section~\ref{logits_shift}) effect of EM.
\end{itemize}
\vspace{0.005mm}
\section{Method}
\subsection{Entropy Minimization Algorithm}
Let $\mathcal{V}$ denote the vocabulary of a pretrained autoregressive language model $p_\theta$, parameterized by $\theta$. Given an input prompt $x$ (e.g., a question or problem description), the model autoregressively generates a response sequence $y = (y_1, y_2, \dots, y_T)$ according to its current policy:

\[
p_\theta(y \mid x) = \prod_{t=1}^{T} p_\theta(y_t \mid y_{<t}, x),
\]

where $T$ is the length of the generated sequence.

Our core idea is to reduce the model’s uncertainty over its own predictions by minimizing the token-level entropy at each generation step. The conditional entropy at time step $t$ is defined as:

\[
H_t = -\sum_{v \in \mathcal{V}} p_\theta(v \mid y_{<t}, x) \log p_\theta(v \mid y_{<t}, x).
\]

To avoid penalizing the prompt portion, we compute entropy only over the generated tokens. Let $T_{\text{prompt}}$ denote the number of tokens in the prompt $x$. Then the set of target positions is:

\[
\mathcal{I} = \{ t \mid t > T_{\text{prompt}},\ y_t \neq \texttt{[PAD]} \}.
\]

The overall EM loss for a single input $x$ is given by:

\[
\mathcal{L}_{\mathrm{EM}}(x;\theta) = \frac{1}{|\mathcal{I}|} \sum_{t \in \mathcal{I}} H_t.
\]

This loss encourages the model to become more confident in its own predictions without relying on external supervision.Entropy Minimization loss is fully differentiable with respect to model parameters, with gradients resembling the score-function estimator found in entropy-regularized reinforcement learning. What's more, Entropy Minimization offers a closed-form objective, eliminating the need for external reward estimation or value baselines, thereby simplifying optimization while retaining the effectiveness of entropy-driven exploration and exploitation.





\subsection{Data Selection}

Entropy minimization (EM) relies on the premise that the model's predictive uncertainty can serve as a meaningful training signal. However, not all input prompts are equally informative in this regard. As noted in prior work~\cite{teacher,1shot}, certain prompts elicit deterministic behavior from the model (e.g., always correct or always incorrect), yielding limited gradient information under entropy-based objectives.

To address this, we adopt a variance-based data selection strategy. Specifically, we measure the variance of the model's \emph{pass@k} accuracy across multiple samples and select prompts for which the model exhibits the highest behavioral variance. This targets inputs on the cusp of the model's capability—neither trivial nor impossible—making them ideal for entropy-driven optimization.

Given a prompt $x$, we draw $k$ independent samples from the model:

\[
\mathcal{Y}^{(x)} = \left\{ y^{(1)}, y^{(2)}, \dots, y^{(k)} \right\}, \quad y^{(i)} \sim p_\theta(\cdot \mid x).
\]

We then compute the \emph{pass@k} score as:

\[
\text{pass@k}(x) = \frac{1}{k} \sum_{i=1}^k \mathbb{I}\left[ y^{(i)} \text{ is correct} \right],
\]

where $\mathbb{I}[\cdot]$ is the indicator function for whether a sample is considered correct (via execution or string match).

We further compute the sample variance of this binary success variable:

\[
\mathrm{Var}_{\text{pass@k}}(x) = \frac{1}{k} \sum_{i=1}^{k} 
    \left(\mathbb{I}\left[ y^{(i)} \text{ is correct} \right] - \text{pass@k}(x) \right)^2.
\]

This variance quantifies the inconsistency of the model's predictions for a given input. A low variance indicates either high confidence in correctness (near-perfect success) or high confidence in failure (uniformly wrong), both of which are suboptimal for entropy minimization, as they lead to low-entropy posteriors that cannot be further improved.

We therefore define our data selection objective as:

\[
x^* = \arg\max_{x \in \mathcal{D}} \mathrm{Var}_{\text{pass@k}}(x),
\]

where $\mathcal{D}$ denotes the unlabeled data pool.

This approach effectively prioritizes those prompts where the model exhibits the most behavioral uncertainty, making them ``entropy-sensitive.'' Such prompts are empirically found to produce the largest entropy gradients and hence drive meaningful parameter updates under EM.

Intuitively, data with high \emph{pass@k} variance suggests that the model's response distribution is straddling the decision boundary—sometimes correct, sometimes not—indicating a broad or multimodal predictive distribution. These are precisely the regions where entropy minimization is most impactful: it encourages the model to concentrate its probability mass on a consistent and (ideally) correct reasoning trajectory.

By contrast, if a model consistently answers a question correctly or incorrectly regardless of sampling, the entropy is either already minimal or optimization is ineffective. Thus, high-variance prompts provide the richest signal for improving model calibration and reasoning fidelity.

A sample from the NuminaMath\cite{numina} dataset that meets the data filtering criteria is as follows:
\vspace{2mm}

\begin{tcolorbox}[
    colframe=blue!70!black, 
    colback=blue!10!white, 
    coltitle=white, 
    fonttitle=\bfseries, 
    title=An example of a selected data\label{long_open_q}, 
    sharp corners, 
    boxrule=0.5mm, 
]
\textbf{Problem}:  
The pressure \( P \) exerted by wind on a sail varies jointly as the area \( A \) of the sail and the cube of the wind’s velocity \( V \). When the velocity is \( 8 \) miles per hour, the pressure on a sail of \( 2 \) square feet is \( 4 \) pounds. Find the wind velocity when the pressure on \( 4 \) square feet of sail is \( 32 \) pounds. \\

\textbf{Solution}:  
12.8
\label{puzzle}
\end{tcolorbox}
\section{Experiment}
\label{main}

\subsection{Experimental Setting}


We implemented the overall training process of EM based on Acclerate~\cite{accelerate}. We selected 1 piece of data from the dataset as prompt. Since it is an unsupervised method, we do not need any data labels. We directly train the model for only 10 steps with a constant learning rate of $2 \times 10^{-5}$, a temperature of 0.5, and a batch size of 64. The reason why only 10 steps are sufficient will be explained in detail in Section~\ref{loss_avg}.

\begin{table}[H]
    \centering
    \small
    \renewcommand{\arraystretch}{1.6}
    \Large
    \resizebox{\linewidth}{!}{
    \begin{tabular}{@{}l *{9}{c} @{}}
    \toprule
\textbf{Model} 
  & \makecell{\textbf{Dataset} \\ \textbf{Size}} 
  & \makecell{\textbf{Training} \\ \textbf{Step}} 
  & \makecell{\textbf{MATH} \\ \textbf{500}} 
  & \makecell{\textbf{Minerva} \\ \textbf{Math}} 
  & \makecell{\textbf{Olympiad} \\ \textbf{Bench}} 
  & \makecell{\textbf{AMC23}} 
  & \makecell{\textbf{KK}} 
  & \makecell{\textbf{MBPP}} 
  & \makecell{\textbf{Avg.}} \\
    \midrule
        \textbf{OpenReasoner-Zero-7B}\cite{orz} & 129k & 600+ & 79.2 & 31.6 & 44.0 & 47.0 & 27.2 & 78.3 & 51.1  \\ 
        \textbf{SimpleRL-Zoo}\cite{zoo} & 24K & 4000 & 76.8 & 30.9 & 39.4 & 55.3 & 17.4 & 75.9 & 49.3 \\ 
        \textbf{Prime-Zero-7B}\cite{prime} & 230K & 240 & 83.8 & 36.0 & 40.9 & 62.7 & 4.4 & 41.0 & 44.8 \\ 
        \textbf{Oat-Zero-7B}\cite{under} & 12K & 300 & 80.0 & 30.1 & 41.0 & 62.7 & 2.2 & 58.7 & 45.8 \\ 
        \midrule
        \textbf{RLVR}\cite{1shot} & 1.2 K & 1000 & 78.6 & 33.8 & 41.6 & 62.5 & 12.4 & 54.5 & 47.2 \\ 
        \textbf{RLVR 16-shot} & 16 & 1000 & 77.8 & 35.3 & 39.9 & 62.2 & 2.4 & 58.5 & 46.0\\ 
        \textbf{RLVR 1-shot} & 1 & 1000 & 78.6 & 36.0 & 43.7 & 61.9 & 5.8 & 60.3 & 47.7 \\ 
        \midrule
        \textbf{Qwen2.5-Math-7B} & NA & NA & 53.0 & 11.0 & 17.2 & 44.1 & 1.0 & 48.9 & 29.2 \\ 
    \rowcolor{bg!70}
    + \textbf{EM 1-shot}& \textbf{1} & \textbf{10}  & 78.8\bbonus{25.8} & 35.3\bbonus{24.3} & 39.7\bbonus{22.5} & 70.3\bbonus{26.2} & 17.4\bbonus{16.4} & 65.1\bbonus{16.2} & 51.1\bbonus{24.9}\\

    \bottomrule
    \end{tabular}
    }
    \vspace{4mm}
    \caption{\centering Comparison of different methods on math reasoning benchmarks (MATH500\cite{math500}, MinervaMath\cite{minerva}, OlympiadBench\cite{olympiad}, AMC23), logic reasoning benchmark (KK\cite{kk}), and code benchmark (MBPP\cite{mbpp}). To reduce randomness, each benchmark is evaluated under avg@8.}
    \label{tab:main}
\end{table}

\subsection{Main Result}

We present our experimental results in Table~\ref{main}. Compared with most RL-based baselines, our EM 1-shot results show strong competitiveness. Specifically, when applying our 1-shot EM method to the Qwen2.5-Math-7B base model, substantial performance improvements are observed across all evaluated math reasoning benchmarks. The performance on MATH500 significantly increases by 25.8 points (from 53.0 to 78.8), and similarly large improvements occur in Minerva Math (+24.3 points), Olympiad Bench (+22.5 points), and AMC23 (+26.2 points). On average, the EM 1-shot strategy achieves an impressive gain of 24.7 points compared to the original Qwen2.5-Math-7B model.

Notably, even with only a single-shot example and minimal training steps (only 10), EM dramatically reduces the gap between Qwen2.5-Math-7B and state-of-the-art RL-based models such as Prime-Zero-7B and RLVR-GRPO. In particular, on the AMC23 benchmark, the EM-enhanced Qwen2.5-Math-7B achieves a competitive score of 70.3, nearing the leading RL models. These results clearly indicate that Entropy Minimization (EM), despite being a simpler and more data-efficient technique compared to typical reinforcement learning methods, has substantial potential to enhance the performance of foundational language models in math reasoning tasks.

During training, the model repeatedly reinforces its own confidence, causing the logits distribution to gradually shift to the right—a phenomenon that will also be discussed in detail in Section~\ref{logits_shift}.

\subsection{Logits Shift}
\label{logits_shift}
We sample 20 prompts from the NuminaMath\cite{numina} dataset and generate responses using four different models (Qwen2.5-Math-7B, Qwen2.5-Math-7B-EM, Qwen2.5-Math-7B-RL, Qwen2.5-Math-7B-EM-RL). Each model generates 20 responses, resulting in a total of $4 \times 20 = 80$ outputs. For each output, we extract the logits, defined as the unnormalized scores $z_i \in \mathbb{R}$ produced by the model before applying the softmax function.

Given a set of logits $\{z_1, z_2, \dots, z_n\}$, the probability for each token is computed as:
\[
p_i = \frac{e^{z_i}}{\sum_{j=1}^{n} e^{z_j}}.
\]
To analyze the overall distributional behavior, we flatten all collected logits across all responses into a single one-dimensional vector:
\[
\mathbf{z}_{\text{flat}} = \{ z_1^{(1)}, z_2^{(1)}, \dots, z_{m_1}^{(1)}, z_1^{(2)}, \dots, z_{m_2}^{(2)}, \dots, z_{m_k}^{(k)} \},
\]
where $z_{j}^{(i)}$ denotes the $j$-th logit in the $i$-th sample, and $k = 80$ is the total number of generated samples. This flattening allows us to perform global statistical analyses.

To quantify the asymmetry of the flattened logits distribution, we compute the skewness:
\[
\text{Skewness} = \frac{1}{n} \sum_{i=1}^{n} \left( \frac{z_i - \mu}{\sigma} \right)^3,
\]
where $\mu$ and $\sigma$ are the mean and standard deviation of the flattened logits vector $\mathbf{z}_{\text{flat}}$, respectively. A positive skewness indicates a right-tailed distribution, whereas a negative skewness indicates a left-tailed distribution.

\begin{figure}[H]
    \centering
    \includegraphics[width=1\linewidth]{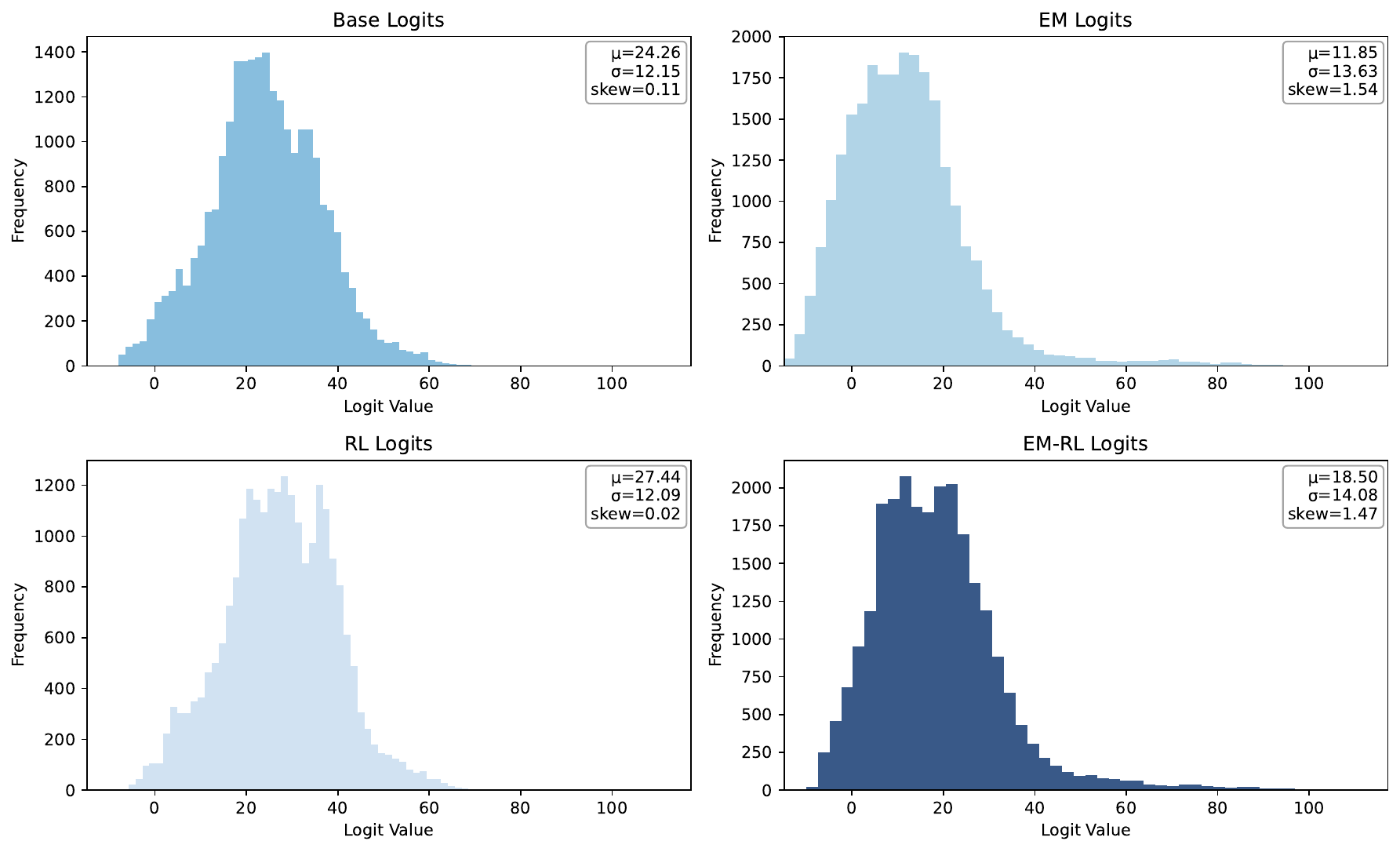}
    \caption{Distribution plot of the flattened logits sampled from 20 sample data points for each of the four models: Base, EM, RL, and RL after EM.}
    \label{logits}
\end{figure}

As observed in Figure~\ref{logits}, models trained with EM exhibit a significantly higher skewness in their logits distribution, indicating a rightward shift. This suggests that EM increases overall model confidence, concentrating probability mass on a subset of tokens. As a result, previously high-probability regions in the original logits become extended into long-tail high-probability intervals.

In contrast, models trained with RL show a pronounced reduction in logits skewness, indicating a leftward shift in the distribution. We hypothesize that this is due to the influence of ground-truth signals during RL training. Specifically, RL appears to perform re-ranking that suppresses tokens with high predicted probability but low alignment with ground truth, thereby reducing their rank and shifting the overall distribution leftward. Similarly, models trained with EM followed by RL exhibit the same pattern: after RL, the skewness of the model's logits decreases from 1.54 (after EM) to 1.47, aligning with this pattern.

We refer to this phenomenon as \textbf{\emph{logits shift}}. A rightward logits shift is beneficial to the generation and sampling process of large language models, as the model primarily samples from high-probability tokens. Shifting the logits to the right increases the number of candidate tokens and expands high-probability paths, thereby enhancing model capability. In contrast, a leftward logits shift is considered harmful to the generation process of large language models, as it reduces the number of high-probability paths during sampling—opposite to the effect of a rightward shift—thus diminishing the model's overall performance. Therefore, the rightward logits shift induced by EM is preferable to the leftward shift caused by RL.

\subsection{Training Loss vs. Reasoning Performance}
\label{loss_avg}

\begin{figure}[H]
    \centering
    \includegraphics[width=0.8\linewidth]{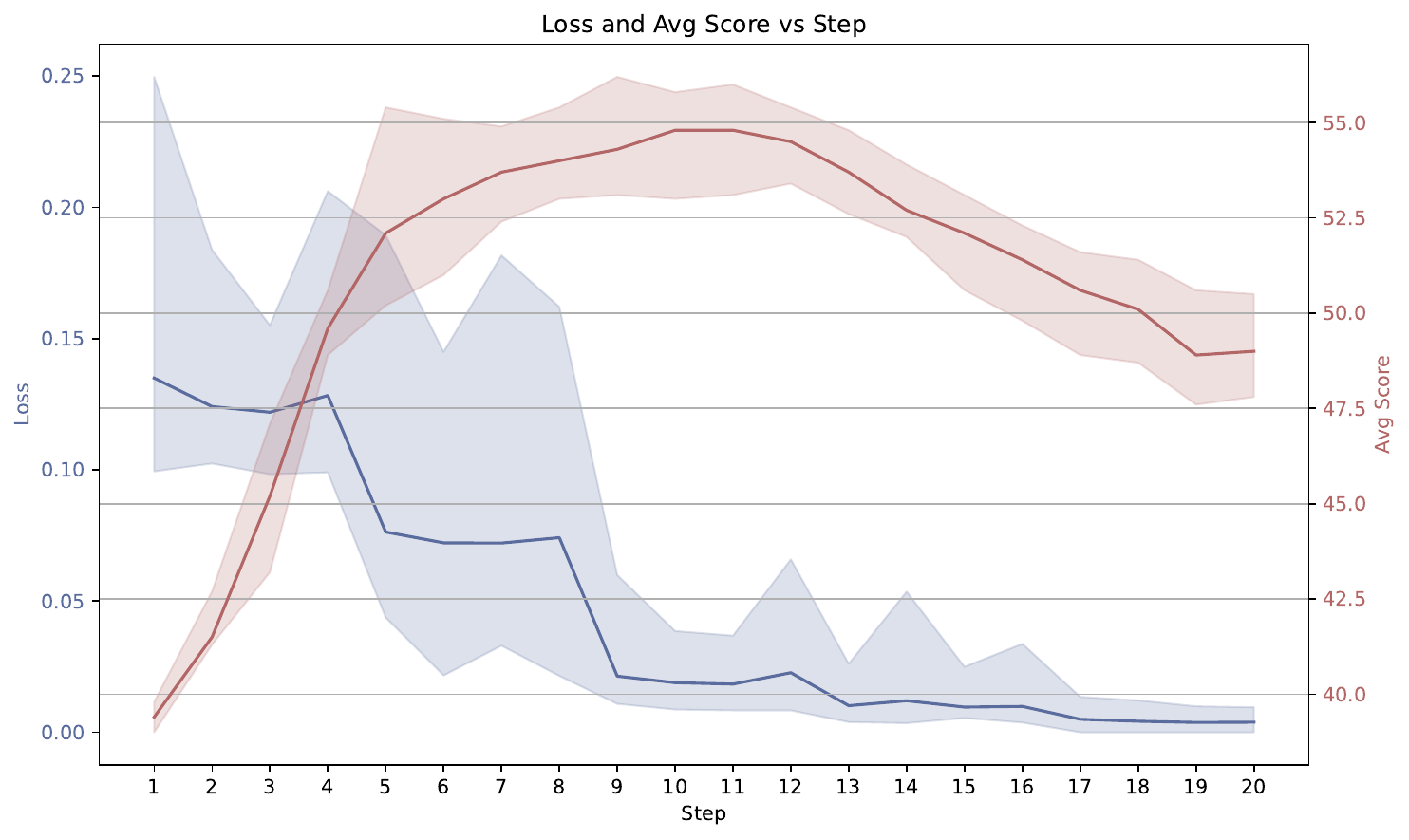}
    \caption{The left y-axis represents the EM training loss, while the right y-axis shows the average score across four reasoning benchmarks. It can be observed that the training loss converges rapidly, whereas the average score peaks around step 10 and then begins to decline. The results in the figure are obtained by repeating the experiments with the same training hyper-parameters using 16 different seeds to reduce randomness.}
    \label{fig:loss_avg}
\end{figure}

As shown in Figure~\ref{fig:loss_avg}, the loss drops to a relatively low level around step 10, and the model's performance on mathematical reasoning reaches its peak. However, unexpectedly, as the EM training loss continues to decrease beyond step 10, the mathematical reasoning performance begins to decline. We designate this phenomenon as \textbf{\textit{over confidence}}. Persistent EM may excessively amplify the model’s confidence in its tokens during inference, thereby exacerbating algorithmic bias and leading to significant deviations in outputs. In conjunction with the findings presented in Section~\ref{logits_shift}, we argue that EM functions primarily as a tool for shaping the model’s distribution rather than as a learning method or strategy. 
Consequently, the effect of distribution shaping is largely achieved within a very small number of training steps, leading to a decoupling between the continued decrease in EM training loss and improvements in mathematical reasoning performance.

\subsection{Sampling Temperature in Training}
\vspace{-2mm}
\begin{figure}
    \centering
    \hspace*{-10mm}
    \includegraphics[width=0.8\linewidth]{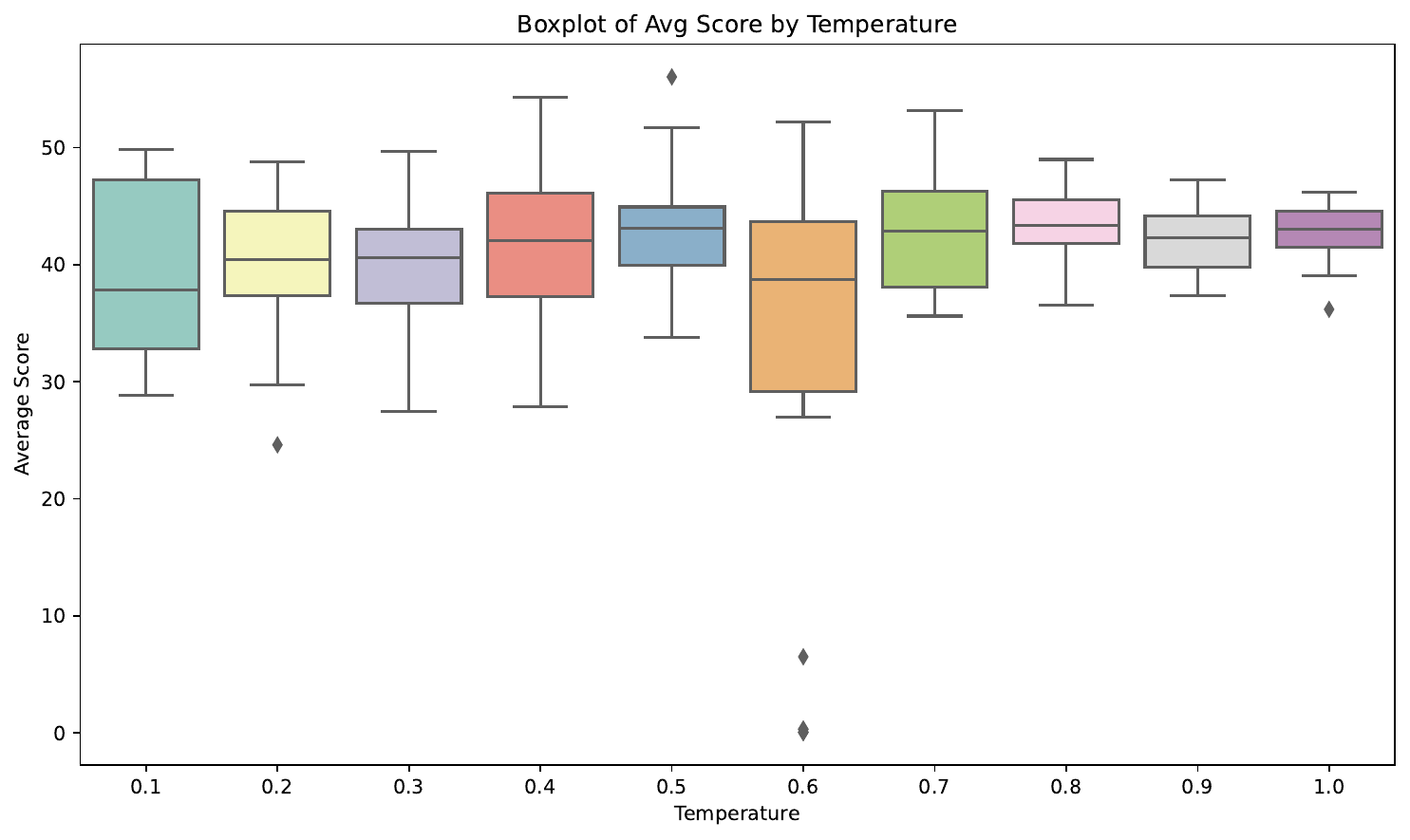}
    \caption{The impact of generation temperature during EM training on the average performance of the trained model across four reasoning datasets. The results in the figure are obtained by repeating the experiments with the same training hyper-parameters using 16 different seeds to reduce randomness.}
    \label{train_temp}
\end{figure}

As shown in Figure~\ref{train_temp}, the average performance of the EM-trained model across four math reasoning benchmarks generally exhibits an upward trend as the generation temperature increases. The maximum of the average performance initially increases and then declines around a temperature of 0.5. Higher temperatures lead to better average reasoning ability, while moderate temperatures (e.g., 0.5) result in greater performance variance, thereby creating opportunities for higher peak performance. Therefore, we prioritize the model trained at temperature 0.5 when reporting final results.

However, as shown in the figure, EM training exhibits significant randomness. The results in the figure are obtained by repeating experiments with 16 different random seeds under the same set of hyperparameters. It can be seen that, even with identical settings, the average scores across the four math reasoning benchmarks can differ by as much as a factor of two depending on the seed. Therefore, all the conclusions in this paper are based on at least 16 repeated experiments with different seeds. We also advocate that future research should focus on reducing the stochasticity of EM training.

\subsection{Sampling Temperature in Evaluate Generation}

\begin{figure}[H]
    \centering
    \includegraphics[width=1\linewidth]{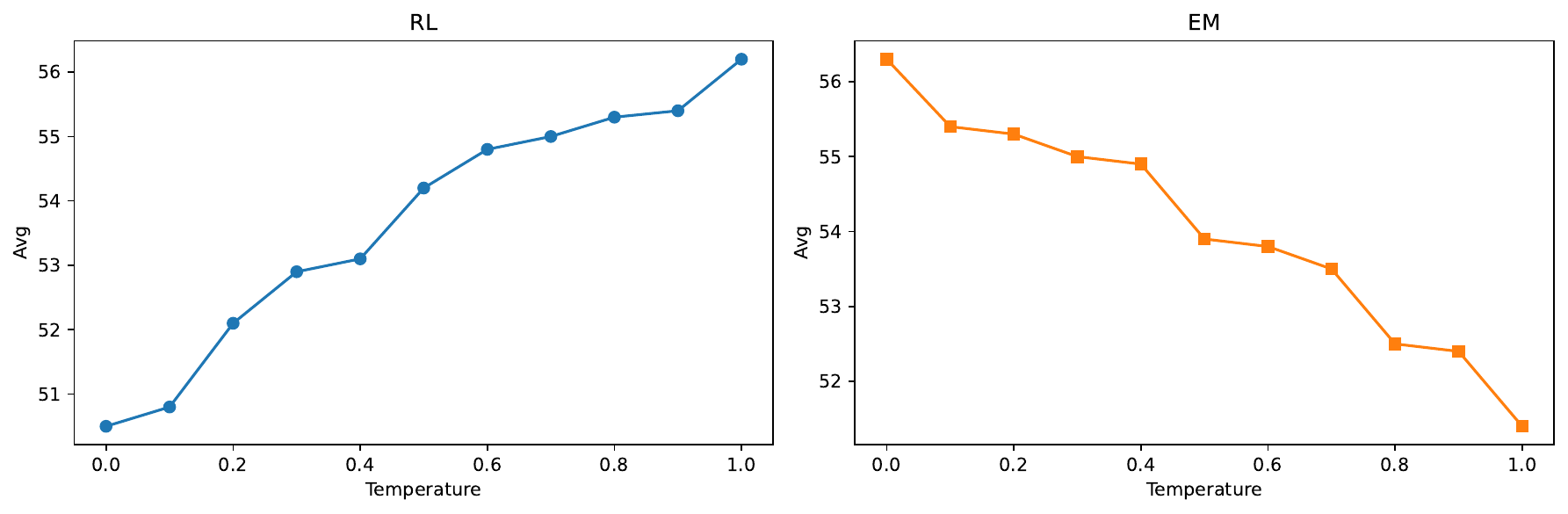}
    \caption{The impact of generation temperature during evalutating on the average performance of
the trained model across four reasoning datasets. The results in the figure are obtained by repeating the
experiments with the same training hyper-parameters using 16 different seeds to reduce randomness.}
    \label{eval_temp}
\end{figure}

As shown in Figure~\ref{eval_temp}, we observe a striking pattern in the EM-trained model's response to varying sampling temperatures during generation. Specifically, as the sampling temperature increases during evaluation, the model’s average performance across four math reasoning benchmarks consistently \textit{decreases}. This trend is in sharp contrast to that of Reinforcement Learning (RL)-trained models shown in Figure~\ref{eval_temp}, where higher sampling temperatures often improve performance.

\paragraph{Greedy Decoding.} The observation can be formally contextualized through the greedy decoding process, which selects the token with maximum conditional probability at each step:

\[
y_t = \arg\max_{v \in \mathcal{V}} \; p_\theta(v \mid y_{<t}, x),
\]

where $\mathcal{V}$ is the vocabulary and $x$ is the input prompt.

Together with our analysis in Section~\ref{logits_shift}, we hypothesize that EM training systematically reshapes the model's logits distributions to become increasingly right-skewed. This process reinforces confidence in already high-probability tokens, effectively concentrating the probability mass on semantically coherent and correct options. As a result, greedy decoding—which deterministically selects the most probable token—becomes a highly effective strategy after EM training.

In contrast, RL adjust token probabilities based on external ground-truth reward. This often promotes the relative ranking of previously low-probability (tail) tokens. Even after reranking, these tokens tend to occupy intermediate positions in the probability distribution, requiring higher temperatures during sampling to be selected. Consequently, RL-trained models exhibit the opposite trend: performance improves with higher sampling temperatures, as seen in Figure~\ref{eval_temp}.

\subsection{EM Before/After RL}

\begin{figure}[H]
    \centering
    \includegraphics[width=1\linewidth]{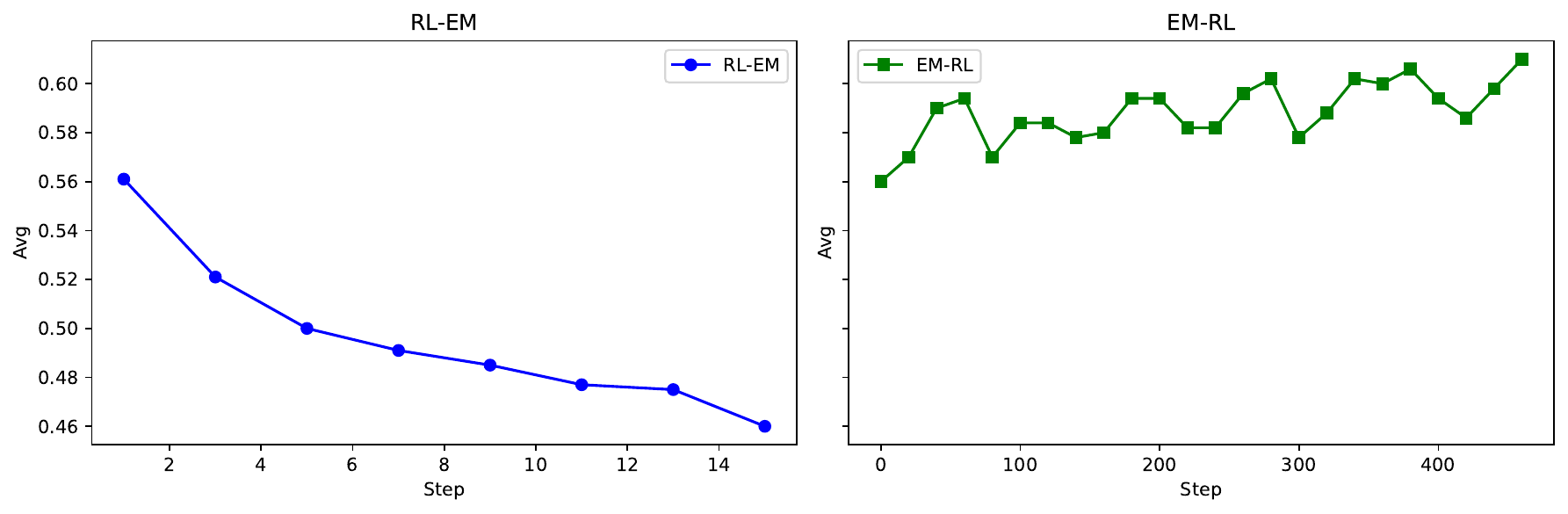}
    \caption{The blue curve on the left represents the average performance across four mathematical reasoning benchmarks of the model trained with RL during the EM phase as training steps progress, while the curve on the right shows the performance of the model trained with EM during the RL phase. The results in the figure are obtained by repeating the experiments with the same training hyper-parameters using 16 different seeds to reduce randomness.}
    \label{rlem}
\end{figure}

Figure~\ref{rlem} shows a clear asymmetry: applying Entropy Minimization (EM) after Reinforcement Learning (RL) leads to a steady decline in performance across four math benchmarks, whereas applying RL after EM yields consistent gains. This suggests that EM exacerbates the distributional distortions introduced by RL, reinforcing the “alignment tax” of RL.

This aligns with prior work showing that RL is most effective when preceded by Supervised Fine-Tuning (SFT)~\cite{open,mpo}, while applying SFT or entropy-based methods after RL often harms performance. In our case, EM after RL locks in narrow, overconfident output modes, while EM before RL enhances reasoning and allows RL to refine outputs without degrading diversity or accuracy.

\subsection{EM on Various Models}

In Table~\ref{tab:model_comparison}, we report the peak performance of 1-shot EM after minimal training steps on various base models. Remarkably, with only a single exemplar and minimal optimization, EM consistently boosts inference accuracy across all models on tasks MATH500, Minerva Math, Olympiad Bench, and AMC23. For instance, on the Qwen2.5-Math-7B model, 1-shot EM yields impressive improvements of 25.8 points on MATH500 (from 53.0 to 78.8), 24.3 points on Minerva Math (from 11.0 to 35.3), 22.5 points on Olympiad Bench (from 17.2 to 39.7), and 26.2 points on AMC23 (from 44.1 to 70.3). This demonstrates EM's ability to substantially enhance reasoning performance with virtually no extra data or compute. Consequently, EM emerges as a compelling lightweight alternative to RL across diverse model bases.

We observe that the upper bound of 1-shot EM’s gains is determined by the base model’s intrinsic reasoning strength. On the relatively weak LLaMA-3.1-8B, 1-shot EM raises average accuracy only to 24.3\%, barely surpassing the baseline of 23.6\%. This suggests that when the underlying model lacks sufficient reasoning capacity, minimal EM optimization cannot fully compensate for its deficits. In contrast, on the stronger Qwen2.5-7B base, 1-shot EM elevates average accuracy significantly from 29.6\% to 37.3\%. Even more notably, on the highly capable Qwen2.5-7B-Instruct base, 1-shot EM pushes average accuracy from 43.12\% to 44.5\%, not only surpassing the baseline but also outperforming EM alone.

Crucially, EM requires no external instruction–response pairs or reward models and can be applied on top of existing SFT or RL checkpoints as a "confidence compression" layer, enabling faster convergence and more stable outputs from very few exemplars. However, interestingly, on the RL-trained SimpleRL-Zoo base, 1-shot EM resulted in a accuracy drop from 44.14\% to 39.1\%, suggesting potential limitations when applied to models already extensively fine-tuned with RL methods. Nonetheless, across most scenarios—especially in SFT and minimally trained RL contexts—EM significantly prunes redundant decision paths and stabilizes key predictions, validating its effectiveness and versatility as a minimal yet powerful optimization strategy.

\subsection{1-shot vs. Multi-shot}
\begin{table}[H]
    \centering
    \small
    \renewcommand{\arraystretch}{1.3}
    \resizebox{1.0\linewidth}{!}{
    \begin{tabular}{l *{7}{c}}
    \toprule
    \textbf{Model}
  & \makecell{\textbf{MATH} \\ \textbf{500}} 
  & \makecell{\textbf{Minerva} \\ \textbf{Math}} 
  & \makecell{\textbf{Olympiad} \\ \textbf{Bench}} 
  & \textbf{AMC23} 
  & \textbf{KK} 
  & \textbf{MBPP} 
  & \textbf{Avg.} \\
    \midrule

    \textbf{LLaMA-3.1-8B} &  49.0 & 21.3 & 15.4 & 25.6 & 7.6 & 73.0 & 32.0 \\ 
    + \textbf{EM multi-shot} & 49.2 & 23.2 & 13.8 & 25.9 & 7.6 & 73.0 & 32.1 \\ 
    + \textbf{EM 1-shot}& 42.6 & 21.0 & 17.0 & 37.8 & 10.2 & 70.1 & 33.1 \\ 
    \midrule
    \textbf{Qwen2.5-7B} & 59.8 & 10.3 & 29.2 & 40.3 & 11.8 & 72.5 & 37.3 \\ 
    + \textbf{EM multi-shot} & 61.8 & 18.4 & 31.1 & 43.4 & 14.0 & 78.3 & 41.2 \\ 
    + \textbf{EM 1-shot} & 67.4 & 22.1 & 33.6 & 45.6 & 12.6 & 75.9 & 42.9 \\
    \midrule
    \textbf{Qwen2.5-7B-Instruct} & 77.0 & 37.1 & 37.9 & 51.9 & 17.0 & 80.7 & 50.3 \\ 
    + \textbf{EM multi-shot} & 74.6 & 34.2 & 36.4 & 48.4 & 11.0 & 78.3 & 47.2 \\  
    + \textbf{EM 1-shot} & 76.4 & 36.4 & 39.3 & 54.1 & 14.8 & 77.5 & 49.8 \\ 
    \midrule
    \textbf{SimpleRL-Zoo} &  76.8 & 30.9 & 39.4 & 55.3 & 17.4 & 75.9 & 49.3 \\ 
    + \textbf{EM multi-shot} & 77.2 & 29.0 & 35.7 & 55.9 & 17.8 & 73.8 & 48.2 \\ 
    + \textbf{EM 1-shot} & 73.0 & 24.6 & 35.3 & 45.9 & 15.4 & 73.0 & 44.5 \\ 
    \midrule
    \textbf{Qwen2.5-Math-7B} & 53.0 & 11.0 & 17.2 & 44.1 & 1.0 & 48.9 & 29.2 \\ 
    + \textbf{EM multi-shot} & 68.0 & 18.8 & 33.2 & 61.6 & 9.0 & 67.2 & 43.0 \\ 
    + \textbf{EM 1-shot} & 78.8 & 35.3 & 39.7 & 70.3 & 17.4 & 65.1 & 51.1 \\
    \bottomrule
    \end{tabular}
    }
    \vspace{7mm}
    \caption{\centering Comparison of different methods on math reasoning benchmarks (MATH500\cite{math500}, MinervaMath\cite{minerva}, OlympiadBench\cite{olympiad}, AMC23), logic reasoning benchmark (KK\cite{kk}), and code benchmark (MBPP\cite{mbpp}). To reduce randomness, each benchmark is evaluated under avg@8.}
    \label{tab:model_comparison}
\end{table}

As shown in Table~\ref{tab:model_comparison}, EM training with just one or two examples achieves an average score on four math benchmarks that is on par with a dataset comprising thousands of examples, and even yields improvements, respectively. Data in Table~\ref{tab:model_comparison} further demonstrate that EM using a minimal exemplar set delivers superior generalization and stronger performance across diverse downstream tasks.

To uncover the underlying factors, we conducted a detailed analysis of the 1-shot EM training dynamics. We observed that both prompt length and generated output length remain markedly more stable under 1-shot EM. Moreover, whereas multi-shot EM losses continue to fluctuate significantly after step 3, the 1-shot EM loss steadily declines from step 3 onward and remains at a low level beyond step 10. This indicates that relying on a single exemplar substantially reduces sample bias and narrows output variance, enabling more fine-grained and stable optimization.

Furthermore, as depicted in Figure~\ref{loss_avg}, EM exhibits RL-like post-saturation generalization behavior: the 1-shot EM loss saturates at step 10, highlighting its remarkably fast convergence—just one example is sufficient to achieve significant gains.

Notably, in the late stage of training, EM loss and model performance become decoupled—further loss reduction does not translate into performance gains and may even incur a slight degradation. We attribute this phenomenon to EM’s "over-confidence" effect, wherein the model amplifies its own prior biases, ultimately leading to a performance penalty.

\subsection{Training Learning Rate}
\begin{figure}[H]
    \centering
    \hspace*{-5mm}
    \includegraphics[width=0.8\linewidth]{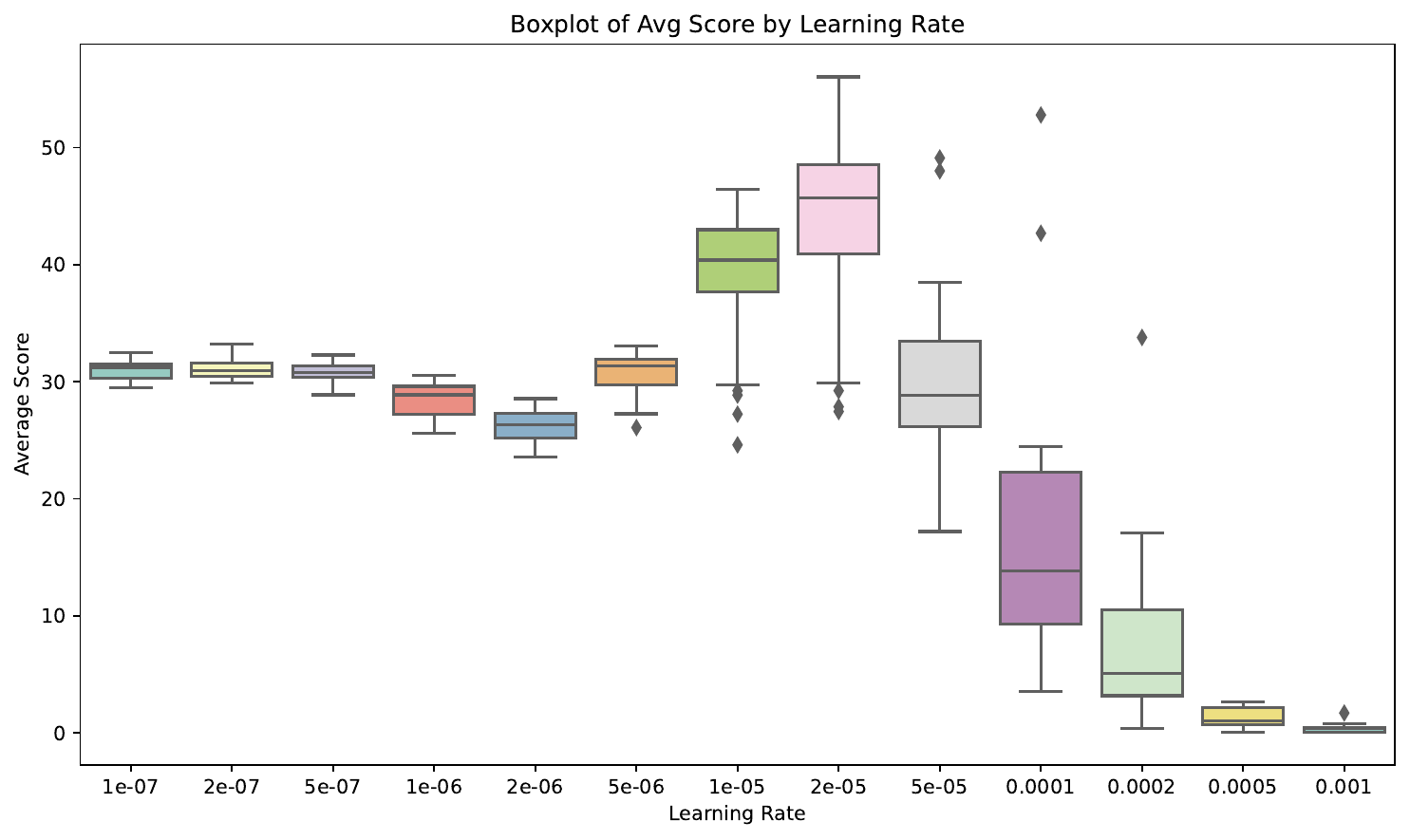}
    \caption{The impact of learning rate during EM training on the average performance of the model at step 10 across four reasoning datasets. The results in the figure are obtained by repeating the experiments with the same training hyper-parameters using 16 different seeds to reduce randomness.}
    \label{lr}
\end{figure}

As shown in Figure~\ref{lr}, a learning rate of $2 \times 10^{-5}$ demonstrates a significant advantage. Since EM typically converges and reaches peak performance within just 10 steps, an excessively large learning rate may cause severe and rapid overconfidence, while a smaller learning rate can slow down convergence. To improve training efficiency, we therefore recommend using $2 \times 10^{-5}$ as the default baseline setting for EM training.

\section{Future Work}

\paragraph{Stablize EM Training.}
One notable finding in our study is the extreme efficiency of entropy minimization (EM), with meaningful improvements achieved in as few as 10 training steps. However, this rapid convergence comes with a potential trade-off: sensitivity to hyperparameters and training instability. Our analysis in Section 3.2 shows that beyond a certain point, continued EM loss reduction may actually harm model reasoning performance. This indicates that EM functions more as a distribution shaping mechanism than a standard learning algorithm—once the model becomes too overconfident, its outputs may collapse into overly narrow token distributions, reducing diversity and correctness. Future work could investigate early stopping criteria or adaptive scheduling mechanisms to stabilize EM training and prevent performance degradation.

\paragraph{Exploring the Full Potential of EM}
Despite its simplicity, EM exhibits surprisingly strong performance with only a single unlabeled example. This raises intriguing questions about how far entropy-based objectives can go. For instance, can EM generalize beyond reasoning tasks to other domains like dialogue, summarization, or code generation? Moreover, current EM setups operate at the token level—future extensions might consider structured entropy over full sequences or semantic units to better capture high-level uncertainty. Incorporating task-specific priors or adaptive entropy regularization may also help unlock further potential from EM training.

Additionally, the connection between entropy minimization and implicit confidence calibration warrants deeper investigation. Our findings suggest that EM enhances model confidence by reinforcing high-probability reasoning paths~\ref{logits_shift}. This implies EM might serve as a lightweight alternative to complex calibration techniques, especially for tasks where interpretability and robustness are critical. Developing evaluation protocols to more precisely quantify EM’s calibration effects will be an important direction.

\paragraph{Combining EM with Other Post-Training Techniques}
Entropy minimization is conceptually orthogonal to most existing post-training paradigms, including supervised fine-tuning (SFT) and reinforcement learning (RL). This opens exciting opportunities for hybrid methods. For example, EM can be applied before SFT to sharpen a model’s predictive distributions, improving its receptiveness to downstream supervision. Alternatively, EM could serve as a regularization strategy during SFT or RLHF.

We also note that applying EM before RL, as briefly explored in Section 3.5, leads to beneficial shifts in logit distributions that may facilitate faster and more stable policy optimization. A systematic study of different EM+RL schedules, curriculum strategies, and interaction effects would help elucidate the best ways to integrate these methods.

\section{Related Work}
\subsection{Entropy Minimization}
Wang et al. \cite{1shot} discovered that RLVR can achieve performance comparable to using thousands of data with just a single data, which is one of the main inspirations for our work. They also were the first to observe that simply optimizing the entropy loss can significantly improve reasoning performance. However, they did not explore this phenomenon in depth, stating in the original paper, “We leave the rigorous analysis to future works.” Agarwal et al. \cite{unreasonable} is the first to study entropy minimization during post-training for large language models. However, they considered the effectiveness of entropy minimization to be unreasonable; although they presented experimental results, their analysis of entropy minimization remained limited.

\subsection{Reinforcement Learning for LLM}
Recent research has increasingly explored post-training approaches to improve the reasoning abilities of large language models. These approaches often involve additional fine-tuning or reinforcement learning using curated datasets that include reasoning tasks and chain-of-thought annotations~\cite{xu2025redstardoesscalinglongcot,logicrl,sc-mcts,zoo,orz,prime,1shot,unreasonable}. Reinforcement learning techniques such as Direct Preference Optimization (DPO)\cite{dpo}, Proximal Policy Optimization (PPO)\cite{ppo}, Group Relative Policy Optimization (GRPO)\cite{grpo}, and REINFORCE++\cite{rpp} are gaining prominence in this area. 
\section{Conclusion}
This work introduces one-shot entropy minimization as a simple yet powerful post-training method for large language models. Using just a single unlabeled example, our approach achieves performance comparable to or better than reinforcement learning methods that rely on large-scale supervision and carefully designed rewards. It is fully unsupervised, highly efficient, and converges within a few training steps. Our experiments show that entropy minimization reshapes the model’s output distribution to increase confidence in correct reasoning paths, effectively enhancing the utility of pretrained knowledge. We identify key indicators—such as logit skewness and behavioral variance—that help target entropy-sensitive inputs and guide optimization.

These findings indicate that significant improvements in reasoning performance can be attained by restructuring existing knowledge through confidence calibration, rather than by acquiring additional information. Entropy minimization thus emerges as both a practical technique and a conceptual framework for advancing our understanding of post-training in large language models.

\section{Acknowledgement}
We thank Tian Xie, Zhengmao Ye, Peihao Wu and Derek Li at Ubiquant for their support and insightful discussions.

\bibliography{references}{}
\bibliographystyle{plain}

\renewcommand{\thesubsection}{\Alph{subsection}}
\end{document}